\documentclass{article}

\PassOptionsToPackage{numbers, compress}{natbib}
\usepackage[preprint]{neurips_2026}

\usepackage[utf8]{inputenc}
\usepackage[T1]{fontenc}
\usepackage{hyperref}
\usepackage{url}
\usepackage{microtype}
\usepackage{graphicx}
\usepackage{subcaption}
\usepackage{multirow}
\usepackage{makecell}
\usepackage{booktabs}
\usepackage{array}
\usepackage[table]{xcolor}
\usepackage{enumitem}
\usepackage{amsmath}
\usepackage{amssymb}
\usepackage{amsfonts}
\usepackage{mathtools}
\usepackage{nicefrac}
\usepackage{algorithm}
\usepackage{algpseudocode}
\usepackage{tikz}
\usetikzlibrary{shapes.geometric,arrows.meta,positioning,fit,calc,
  decorations.pathreplacing,backgrounds,patterns}
\usepackage[font=small,labelfont=bf]{caption}
\usepackage{wrapfig}
\usepackage{pifont}
\usepackage{marvosym}

\newcommand{\ours}{\textsc{DeltaVid}}
\newcommand{\trainset}{\textsc{DeltaVid-10K}}
\newcommand{\mypara}[1]{\vspace{4pt}\noindent\textbf{#1}}
\definecolor{scolor}{HTML}{2E86C1}
\definecolor{tcolor}{HTML}{E67E22}
\definecolor{stcolor}{HTML}{8E44AD}
\definecolor{gcolor}{HTML}{27AE60}

\title{\ours{}: Enhancing Fine-Grained Spatiotemporal Perception\\with Cross-Video Differences}

\makeatletter
\newcommand{\blfootnotetext}[1]{%
  \begingroup
    \renewcommand{\thefootnote}{}%
    \renewcommand{\@makefnmark}{}%
    \long\def\@makefntext##1{\noindent##1}%
    \footnotetext{#1}%
  \endgroup
}
\renewcommand{\@noticestring}{}
\makeatother

\author{%
  \begin{minipage}[t]{\textwidth}\centering
    Yankai Yang\textsuperscript{*\,1,2} \quad
    Yancheng Long\textsuperscript{*\,1,2} \quad
    Bin Wen\textsuperscript{\textdagger\,\Letter\,2} \quad
    Fan Yang\textsuperscript{2} \\[3pt]
    Tingting Gao\textsuperscript{2} \quad
    Han Li\textsuperscript{2} \quad
    Shuo Yang\textsuperscript{\Letter\,1}
  \end{minipage}%
}
\begin{document}
\raggedbottom

\maketitle
\blfootnotetext{%
  \textsuperscript{*}Equal contribution.\quad
  \textsuperscript{\textdagger}Project leader.\quad
  \textsuperscript{\Letter}Corresponding authors.\\[1pt]
  \textsuperscript{1}Harbin Institute of Technology, Shenzhen.\quad
  \textsuperscript{2}Kuaishou Technology.\\[1pt]
  Correspondence to: Shuo Yang $<$shuoyang@hit.edu.cn$>$, Bin Wen $<$wenbin@kuaishou.com$>$.%
}

\begin{abstract}
Video multimodal large language models have made strong progress on open-ended video understanding, but they still lack precise local spatiotemporal perception. When two videos share almost the same global semantics and differ only in a short time span or a small region, current models often fail to find the change and provide reliable evidence. We propose \ours{}, a verifiable proxy-task framework that enhances fine-grained spatiotemporal perception with cross-video differences. The key idea is to turn cross-video spot-the-difference into a trainable perception signal, where a model identifies local changes, judges temporal boundaries, and organizes spatial evidence by comparing similar videos. To make this signal scalable to train and reliable to evaluate, we further introduce \trainset{} and \ours{}-Bench, which convert controllable local differences in real videos into evidence-labeled training and test samples. Experiments show that \ours{} substantially improves performance on cross-video difference understanding and transfers the learned local evidence ability to general video understanding benchmarks, including MMVU, MLVU, Video-MME, VideoHolmes, VideoMMMU, LVBench, TempCompass, and LongVideoBench. These results show that cross-video differences are not only an effective way to diagnose fine-grained perception failures, but also a scalable proxy supervision that moves Video MLLMs from coarse semantic understanding toward fine-grained spatiotemporal evidence reasoning.

\end{abstract}

\begin{figure}[t]
\centering
\includegraphics[width=\textwidth]{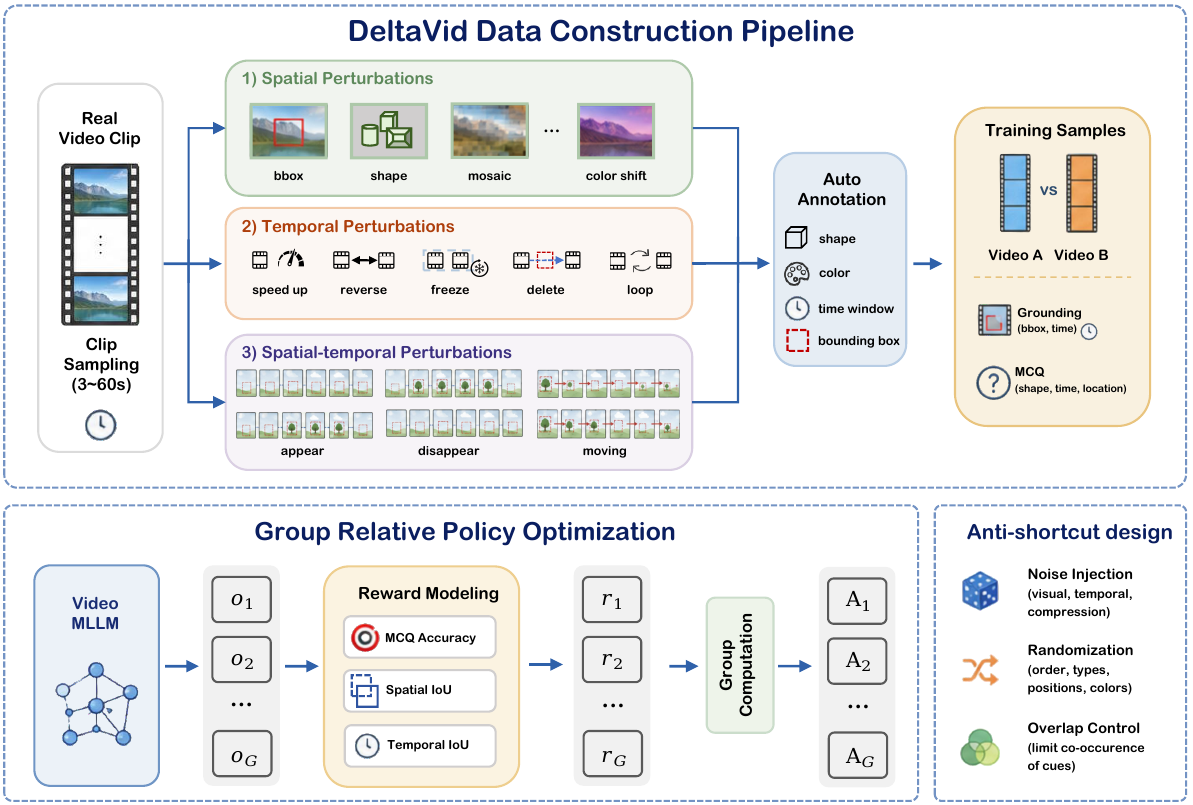}
\caption{\ours{} data construction, rule-reward training, and anti-shortcut design.}
\label{fig:framework}
\end{figure}


\section{Introduction}
\label{sec:intro}

In everyday visual perception, comparing two similar videos and identifying local changes is often intuitive: a human observer can align temporal progression, attend to local regions, and judge whether an object appears, moves, changes color, or changes speed. For video multimodal large language models (Video MLLMs)~\cite{flamingo,blip2,llava,minigpt4,qwen3vl,qwen25vl,llava-video,llava-onevision,internvl2.5,gpt4o}, however, this capability does not emerge naturally. Although these models have made rapid progress in video question answering, long-video captioning, and open-ended reasoning, and can produce fluent global semantic descriptions, they may still miss a small blurred region, a short object appearance, a reversed or sped-up segment, or a local difference that lasts only a few seconds. This suggests that current video pretraining and instruction tuning paradigms can build high-level semantic alignment, but lack direct supervision for local changes, temporal boundaries, and spatial evidence.

Reinforcement learning (RL) post-training has become an important way to improve multimodal models~\cite{deepseekr1,grpo,ppo,dpo,rlhf,visionr1,r1onevision,mmeureka,r1vl,reasonrft}. Recent work shows that rule-based rewards can improve reasoning format, answer quality, and some perception skills without a human preference model. Video methods have also started to explore temporal reasoning and video decision making~\cite{videor1,tempr1,videozoomer}. Still, existing methods often trade task specificity for generalization, or depend on human-labeled data that limits scale and diversity. Direct tuning on standard grounding or QA data can improve task-specific scores, but it may not build fine-grained perception for open video understanding. Proxy tasks such as visual jigsaw puzzles, hallucination detection, and image difference localization~\cite{jigsawr1,vicrit,spotdiff} provide useful verifiable supervision. Yet most of them focus on static images or coarse perception. They do not fully cover temporal boundaries, local motion, and spatiotemporal differences in videos.

To address these limits, we propose \ours{}, a verifiable proxy-task framework for improving fine-grained spatiotemporal perception with cross-video differences. Given two similar videos $V_A$ and $V_B$, the model must identify local changes and output the change type, time span, and/or spatial region according to the task. \ours{} targets video by modeling differences as spatial, temporal, and spatiotemporal local evidence, forcing the model to learn fine-grained cross-video comparison, temporal boundary judgment, and local evidence organization.

Around this task, \ours{} builds an independent training dataset, \trainset{}, from real videos. Controlled spatial, temporal, and spatiotemporal perturbations create high-quality A/B video pairs and provide precise labels for change attributes, time spans, and spatial regions. Training further organizes these labels into two complementary forms of supervision: Grounding teaches the model to point to evidence, while MCQ teaches it to distinguish similar changes. This design lets the model learn local evidence alignment from sparse video difference signals without a learned reward model.

We evaluate \ours{} through a series of experiments. We first use our \ours{}-Bench to directly measure cross-video difference understanding. We then test transfer on standard video understanding benchmarks, including MMVU~\cite{mmvu}, MLVU~\cite{mlvu}, Video-MME~\cite{videomme}, VideoHolmes~\cite{videoholmes}, VideoMMMU~\cite{videommmu}, LVBench~\cite{lvbench}, TempCompass~\cite{tempcompass}, and LongVideoBench~\cite{longvideobench}. The results show that Qwen3-VL-8B trained with \ours{} clearly outperforms strong baselines, including the original Qwen3-VL-8B, Qwen3-VL-235B~\cite{qwen3vl}, and Video-R1~\cite{videor1}, on our difference benchmark. More importantly, the learned fine-grained spatiotemporal skills transfer to general video understanding tasks. These results show that cross-video difference comparison is a scalable and robust proxy task for improving local evidence skills in Video MLLMs. Our main contributions are:

\begin{itemize}[leftmargin=1.5em,itemsep=3pt,topsep=4pt]
\item We introduce and systematically study cross-video difference understanding, showing that existing Video MLLMs still struggle to detect local changes between similar videos, especially for short-lived changes, temporal boundaries, and local spatial evidence alignment.
\item We propose \ours{}, a verifiable proxy-task framework for fine-grained spatiotemporal perception. It constructs fully controllable A/B video pairs from real videos, covers spatial, temporal, and spatiotemporal changes, and automatically produces structured labels for change attributes, time spans, and spatial regions, forming the independent training set \trainset{}.
\item We design a rule-reward post-training strategy that combines Grounding and MCQ, decomposing sparse video difference signals into continuous localization rewards and closed-set attribute discrimination rewards. Experiments show that \ours{} substantially improves the same-scale model on cross-video difference localization, surpasses a larger model and approaches GPT-5, while maintaining stable transfer to general video understanding benchmarks.
\end{itemize}

\section{Related work}
\label{sec:related}

\mypara{Multimodal and video foundation models.}
Vision-language foundation models use image-text alignment and instruction tuning to build a general multimodal paradigm~\cite{flamingo,coca,blip2,instructblip,llava,minigpt4,sharegpt4v}. Video models further handle temporal modeling, long-context compression, and cross-modal memory, including Video-LLaMA~\cite{videollama}, Video-ChatGPT~\cite{videochatgpt}, VideoChat~\cite{videochat}, Video-LLaVA~\cite{videollava}, TimeChat~\cite{timechat}, MovieChat~\cite{moviechat}, LLaMA-VID~\cite{llamavid}, LLaVA-Video~\cite{llava-video}, LLaVA-OneVision~\cite{llava-onevision}, Long-context World Models~\cite{lwm}, and InternVideo2~\cite{internvideo2}. Surveys summarize this fast progress~\cite{videounderstandingsurvey}. These works improve open video interaction and long-video understanding, but their objectives often focus on global semantics, answer correctness, or dialog quality. They rarely give direct supervision for local change evidence.

\mypara{Perception diagnostics and video comparison evaluation.}
Multimodal evaluation has moved from general QA to hallucination, spatial relations, region reference, and integrated skills. Examples include MME~\cite{mme}, MMBench~\cite{mmbench}, MM-Vet~\cite{mmvet}, VSR~\cite{vsr}, HallusionBench~\cite{hallusionbench}, MMVP~\cite{mmvp}, POPE~\cite{pope}, and Cambrian-1~\cite{cambrian1}. Difference-based evaluation also tests fine-grained comparison. Early work studies image spot-the-difference~\cite{spotdiff,clevr}, while VidDiff~\cite{viddiff} and ViDiC~\cite{vidic} extend this idea to video comparison. These works expose perception failures, but they mainly serve as diagnostics or description benchmarks. We instead turn video comparison into scalable post-training supervision.

\mypara{Rule-based post-training and verifiable proxy tasks.}
RL post-training for multimodal models often improves rule following, multi-step reasoning, and answer quality~\cite{deepseekr1,grpo,ppo,dpo,rlhf,visionr1,r1onevision,mmeureka,r1vl,reasonrft}, or optimizes temporal reasoning and video decision skills~\cite{videor1,tempr1,videozoomer}. These works show that rule rewards can shape model behavior without human preference models, but most rewards still target final answers or high-level task correctness. At the same time, Perception-R1~\cite{perceptionr1}, Jigsaw-R1~\cite{jigsawr1}, and ViCrit~\cite{vicrit} show that verifiable proxy tasks can directly improve visual perception. We extend this direction with cross-video difference comparison: the model must find local changes between similar videos, and the supervision covers time spans, spatial regions, and their combination.

\section{Method}
\label{sec:method}

This section describes the training pipeline of \ours{}, as summarized in Figure~\ref{fig:framework}. The key idea is to turn cross-video comparison into verifiable local evidence learning: construct A/B video pairs from real videos, record temporal, spatial, and attribute labels for each controlled difference, convert the labels into Grounding and MCQ tasks, and post-train the model with difficulty-aware data organization, rule-based rewards, and anti-shortcut design. Unlike standard video QA, which mainly supervises final text answers, \ours{} directly constrains the evidence behind the answer.

\subsection{Problem formulation}
\label{sec:formulation}

Given an original video $V_A$, a comparison video $V_B$, and a text prompt $q$, the goal is to predict the differences between the two videos. Each pair contains one or more ground-truth differences $\mathcal{D}^{\star}=\{d_j^{\star}\}_{j=1}^{M}$, where $d_j^{\star}$ is represented by a change type $c_j$, a time span $\tau_j=[t_j^s,t_j^e]$, a spatial box $b_j=[x_1,y_1,x_2,y_2]$, and optional attributes $a_j$. Different differences require different evidence fields: spatial changes require $b_j$, temporal changes require $\tau_j$, and spatiotemporal changes require both.

The model output is either a structured evidence set $\hat{\mathcal{D}}=\{\hat{d}_i\}_{i=1}^{N}$ or an MCQ answer $\hat{o}$. Local change evidence alignment therefore consists of three skills: recognizing the change type, localizing its time span, and localizing its spatial region. Grounding trains temporal and spatial evidence localization, while MCQ trains closed-set attribute discrimination with hard distractors.

\subsection{Difference pair synthesis from real videos}
\label{sec:data}

The training data is built on real videos to preserve natural backgrounds, textures, camera motion, and event dynamics. Controlled perturbations provide exact temporal, spatial, and attribute labels. Given a video collection, we sample 3--60 second clips as Video A and apply one or more local perturbations to form Video B. The underlying pool contains 8,200 A/B video pairs and 10,743 difference instances, covering both single-difference and multi-difference cases. \trainset{} is obtained by converting this pool into training samples.

\mypara{Perturbation taxonomy.}
To cover the core evidence types required by cross-video difference comparison, we construct a compact perturbation set corresponding to three challenges: local spatial localization, temporal boundary detection, and joint spatiotemporal alignment. Table~\ref{tab:perturbation} lists the 13 perturbation types used by \ours{}.

\begin{table}[h]
\centering
\caption{Perturbation taxonomy in \ours{}.}
\label{tab:perturbation}
\small
\begin{tabular}{@{}l l p{6cm}@{}}
\toprule
Dimension & Type & Description \\
\midrule
\multirow{4}{*}{\textcolor{scolor}{\textbf{Spatial}}}
& Shape insertion & Add a colored shape to a local region of Video B during a random time window \\
& Region blur & Apply Gaussian blur to a local region during a random time window \\
& Region pixelation & Pixelate a local region during a random time window \\
& Color shift & Shift the hue of a local region during a random time window \\
\midrule
\multirow{5}{*}{\textcolor{tcolor}{\textbf{Temporal}}}
& Speed change & Speed up or slow down a local video segment \\
& Reverse segment & Reverse the temporal order of a local segment \\
& Freeze frame & Insert a frozen segment at a sampled time \\
& Segment deletion & Delete a local time segment \\
& Segment loop & Repeat a local segment multiple times \\
\midrule
\multirow{4}{*}{\textcolor{stcolor}{\textbf{Spatiotemporal}}}
& Object appears & A shape starts to appear from a sampled time \\
& Object disappears & A shape disappears after a sampled time \\
& Moving object & A shape moves during the video \\
& Timed local blur & Blur a region only during a specific time span \\
\bottomrule
\end{tabular}
\end{table}

\begin{figure}[t]
\centering
\includegraphics[width=\textwidth]{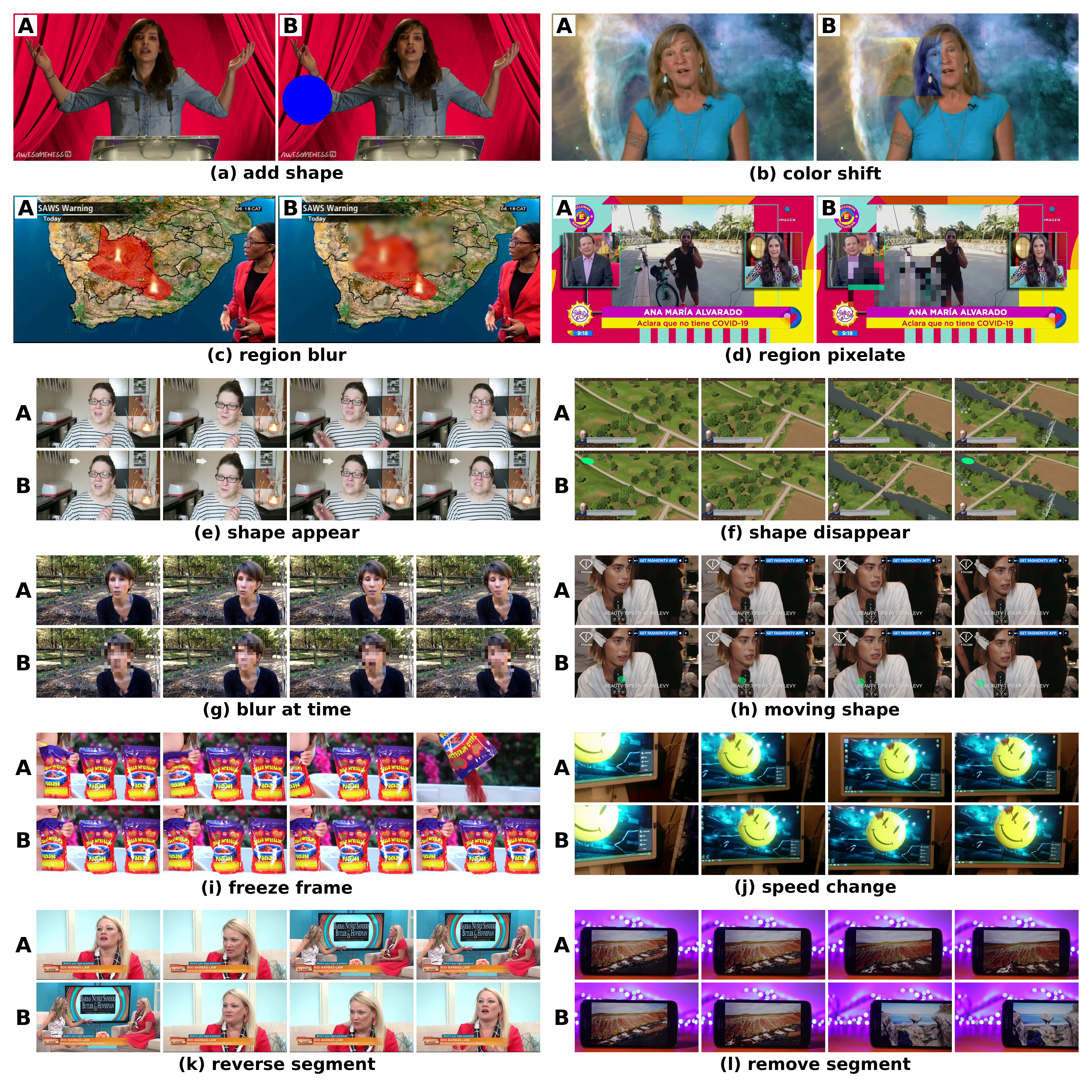}
\caption{Representative perturbation examples in \trainset{}. Each panel shows an A/B video pair or key-frame sequence, covering spatial, spatiotemporal, and temporal perturbation types used to construct training data.}
\label{fig:perturbation_examples}
\end{figure}

Figure~\ref{fig:perturbation_examples} shows representative training examples. \ours{} does not synthesize differences in blank or purely synthetic scenes; instead, it applies local controllable perturbations to real video backgrounds, preserving realistic visual distributions while yielding automatically verifiable spatiotemporal labels.

The taxonomy covers three complementary skills: spatial perturbations require local region localization, temporal perturbations require boundary detection, and spatiotemporal perturbations require joint temporal-spatial alignment. Spatial perturbations are active only within sampled time windows, while temporal and spatiotemporal perturbations directly record temporal boundaries. To reduce shortcut learning, we randomize perturbation parameters, control overlap and ambiguity in multi-difference samples, and use shuffled MCQ options with visually similar distractors. Further sampling details are provided in Appendix~\ref{app:perturbation}.

\subsection{Dual-task generation}
\label{sec:task}

Because the perturbation process records temporal, spatial, and attribute labels for each difference, the same A/B pair can be converted into two complementary tasks.

\mypara{Grounding.}
Grounding converts each difference into verifiable structured evidence. Spatial differences require the model to localize the changed region, temporal differences require the model to predict the occurrence interval, and spatiotemporal differences require both temporal and spatial evidence. Unlike QA supervision that only checks the final answer, Grounding directly constrains the local evidence behind the answer and strengthens fine-grained difference perception.

\mypara{MCQ.}
MCQ provides closed-set supervision for attribute discrimination. The templates cover change type, shape, color, time, location, and speed factor. The main training set uses shape, time, and location questions; options are shuffled, and visually similar distractors are preferred when possible.

\subsection{Difficulty-aware difference training}
\label{sec:difficulty}

Cross-video difference training naturally suffers from sparse rewards. At the beginning of training, a model often fails to simultaneously detect the difference, predict the correct time span, and localize the spatial region; complex multi-difference samples can therefore produce near-zero rewards. To mitigate this issue, \trainset{} is not composed only of the hardest open comparison cases. Instead, it is organized by the number of differences, evidence dimensions, and task forms.

Single-difference samples provide clean local alignment signals and teach the basic mapping of where and when a difference occurs. Dual- and mixed-difference samples require the model to separate multiple local changes, reducing the bias toward reporting only the most salient difference. Spatiotemporal samples further require the model to bind temporal boundaries with spatial regions. This organization does not introduce an additional learned curriculum model; instead, it supplies both basic alignment cases and complex comparison cases during rule-reward training, improving optimization stability.

\subsection{Reinforcement learning training}
\label{sec:training}

\mypara{Reward function.}
The Grounding reward uses temporal IoU and spatial IoU. Spatial samples use bbox IoU, temporal samples use temporal IoU, and spatiotemporal samples use their product:
\begin{equation}
r_{\mathrm{st}} = \mathrm{IoU}_{t}(\hat{\tau}, \tau^\star) \cdot \mathrm{IoU}_{s}(\hat{b}, b^\star).
\end{equation}
The product enforces a joint spatiotemporal constraint: a correct time span with an incorrect region, or a correct region with an incorrect time span, should not receive high reward. For samples with multiple differences, the output order should not become an extra constraint. We therefore select the highest-scoring one-to-one assignment under a bipartite matching constraint; unmatched differences receive zero reward, which penalizes missed changes, duplicate predictions, and coarse localization. MCQ uses exact-match reward:
\begin{equation}
r_{\mathrm{mcq}}=\mathbf{1}[\hat{o}=o^{\star}].
\end{equation}
To make open-ended outputs parseable, we add a format reward. The answer should contain parseable \texttt{<think>...</think>} and \texttt{<answer>...</answer>} fields. The format term has a small weight and maintains structured output; the main optimization signal still comes from temporal, spatial, and MCQ task rewards. The final reward is
\begin{equation}
r = 0.9 r_{\mathrm{task}} + 0.1 r_{\mathrm{fmt}}.
\end{equation}
If parsing fails, the answer receives no task reward. \ours{} uses only structured labels and does not rely on human preference models, model-based scoring, or free-text similarity. The resulting signal is scalable while remaining directly tied to verifiable video evidence.

\mypara{GRPO optimization.}
Following DeepSeekMath~\cite{grpo}, we use Group Relative Policy Optimization (GRPO). For each question, the model samples a group of answers. The reward is normalized by the mean and standard deviation inside the group:
\begin{equation}
\hat{A}_i = \frac{r_i - \mu(\mathbf{r})}{\sigma(\mathbf{r}) + \epsilon}.
\end{equation}
Here $\mathbf{r} = (r_1, \ldots, r_G)$ is the reward group. The three difference dimensions and two task types share one optimization objective; the full pipeline is provided in Appendix Algorithm~\ref{alg:training}.

\section{Experiments}
\label{sec:exp}

We evaluate \ours{} along two axes. First, \ours{}-Bench directly measures whether a model can compare A/B videos and localize the evidence behind the difference. Second, general video benchmarks test whether this local spatiotemporal skill transfers to standard video understanding. Unless otherwise stated, all models use the same video inputs, question formats, and answer parser.

\subsection{Training and data setup}

\mypara{Models.}
We use Qwen3-VL-8B~\cite{qwen3vl} as the base model and post-train it on \trainset{} to obtain \ours{}-Qwen3-VL-8B. Baselines include the original Qwen3-VL-8B, GPT-5, the larger Qwen3-VL-235B~\cite{qwen3vl}, and Video-R1~\cite{videor1}, a video reasoning model trained with RL. These comparisons separate the effect of cross-video difference training from model scale and general video RL training.

\mypara{Data.}
The task composition, input-form distribution, and subset breakdown of \trainset{} are reported in Appendix~\ref{app:data}.

\mypara{Training setup.}
Training uses only \trainset{} with direct GRPO updates. We do not add human preference data or a learned reward model. Except for ablations, all samples are optimized as one training pool. Full hyperparameters, frame sampling settings, and compute details are reported in Appendix~\ref{app:training}.

\subsection{\ours{}-Bench}
\label{sec:vdigbench}

Existing general video benchmarks mainly evaluate single-video QA or broad video understanding, and do not isolate whether a model can compare two highly similar videos and ground the difference in concrete temporal and spatial evidence. We therefore build \ours{}-Bench as a diagnostic benchmark for cross-video difference understanding. Each example contains an A/B video pair and one question, requiring the model to judge the number of differences, change type, temporal order, time span, or spatial region.

\ours{}-Bench contains 500 diagnostic questions organized into seven skill groups, with the composition reported in Appendix Table~\ref{tab:benchcomp}. For evaluation, all models use the same video inputs, question templates, and answer parser. We report MCQ accuracy, temporal localization score, spatial localization score, and an overall score. All samples come from manually curated high-quality video pairs and are verified for answerability, lack of ambiguity, and answer correctness; construction details and agreement statistics are provided in Appendix~\ref{app:bench}.

\subsection{General evaluation setup}
\label{sec:evalsetup}

We further evaluate on nine general video benchmarks: MMVU~\cite{mmvu}, MLVU~\cite{mlvu}, MVBench, Video-MME~\cite{videomme}, VideoHolmes~\cite{videoholmes}, VideoMMMU~\cite{videommmu}, LVBench~\cite{lvbench}, TempCompass~\cite{tempcompass}, and LongVideoBench~\cite{longvideobench}. These benchmarks cover long-video understanding, multidisciplinary video QA, temporal order reasoning, fine-grained event judgment, and long-context video-language understanding, and test whether \ours{} transfers beyond the in-domain difference benchmark.

\subsection{Main results}

\mypara{Cross-video difference comparison on \ours{}-Bench.}
We first evaluate the core cross-video difference capability on \ours{}-Bench, testing whether models can discover, localize, and discriminate local changes from A/B videos. The results are shown in Table~\ref{tab:vdigbench}.

\begin{table}[t]
\centering
\caption{\ours{}-Bench results.}
\label{tab:vdigbench}
\small
\setlength{\tabcolsep}{4pt}
\begin{tabular}{@{}lcccc@{}}
\toprule
Model & MCQ & T-IoU & S-IoU & Overall \\
\midrule
Video-R1~\cite{videor1} & 68.00 & 30.29 & 62.22 & 52.20 \\
Qwen3-VL-8B & 65.00 & 40.57 & 69.78 & 58.60 \\
Qwen3-VL-235B & 72.00 & 58.86 & 67.11 & 65.20 \\
GPT-5 & \textbf{78.00} & 55.43 & \textbf{77.78} & \textbf{70.00} \\
\textbf{\ours{}-Qwen3-VL-8B} & 69.00 & \textbf{64.00} & 69.33 & \emph{67.40} \\
\bottomrule
\end{tabular}
\end{table}

Table~\ref{tab:vdigbench} shows two findings. First, existing video models still struggle with cross-video spot-the-difference, especially when localizing short-lived differences. Second, after \ours{} training, the 8B model improves substantially over the original Qwen3-VL-8B, surpasses the larger Qwen3-VL-235B overall, and approaches GPT-5-level performance. Per-metric results show that the main gain comes from temporal evidence localization: T-IoU improves most over the base model and is the highest among all compared models. At the same time, MCQ improves and S-IoU remains largely stable, suggesting that the training signal preserves spatial evidence grounding while specifically strengthening the more difficult temporal boundary modeling required by cross-video differences.

\begin{figure}[t]
\centering
\includegraphics[width=\textwidth]{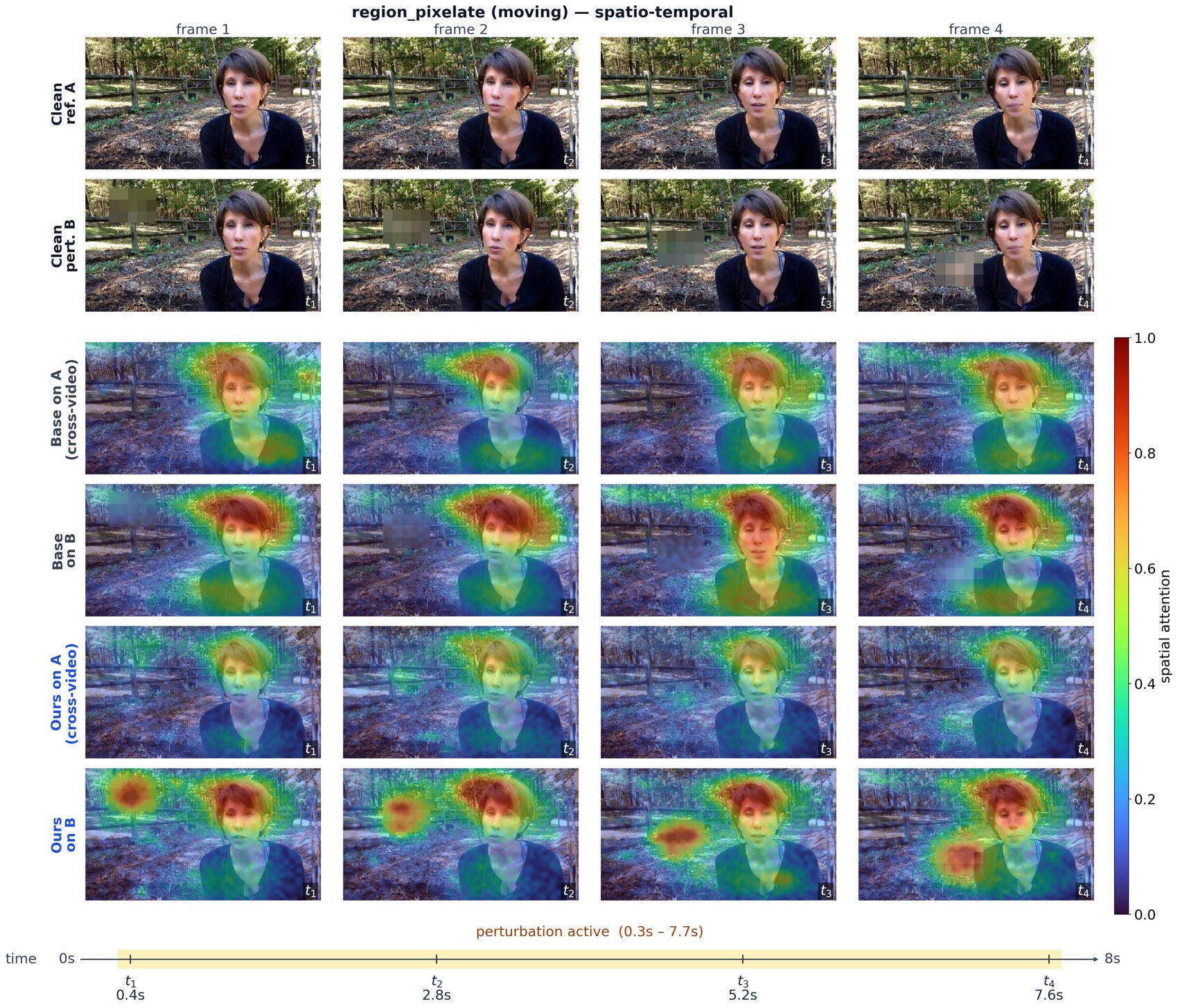}
\caption{Attention visualization example. The first two rows show key frames from the reference Video A and perturbed Video B; the middle two rows show spatial attention from the base model under cross-video input; the bottom two rows show attention after \ours{} training. Compared with the base model, \ours{} more consistently attends to the local changed region in the perturbed video.}
\label{fig:attention_case}
\end{figure}

To further inspect whether the model attends to difference evidence, Figure~\ref{fig:attention_case} shows an attention visualization example. The base model mainly attends to the human subject and globally salient regions, whereas the model trained with \ours{} focuses more consistently on the local changed region in Video B across frames. This observation is consistent with the T-IoU improvement in Table~\ref{tab:vdigbench}, suggesting that cross-video difference training helps bind model decisions to concrete spatiotemporal evidence.

\mypara{Transfer to general video understanding.}
We next test whether the local spatiotemporal evidence skill learned by \ours{} transfers to standard video understanding. Unlike \ours{}-Bench, these benchmarks do not require explicit A/B comparison, so they test whether cross-video difference training merely fits the in-domain task or induces more general temporal and spatial perception.

\begin{table}[t]
\centering
\caption{General video understanding results.}
\label{tab:general}
\small
\setlength{\tabcolsep}{5pt}
\begin{tabular}{@{}lccccc@{}}
\toprule
Model & MMVU & MLVU & Video-MME & VideoHolmes & VideoMMMU \\
\midrule
Video-R1~\cite{videor1} & 63.80 & -- & 61.40 & -- & 52.40 \\
Qwen3-VL-8B & 60.80 & 74.20 & 66.89 & 43.60 & 57.00 \\
\textbf{\ours{}-Qwen3-VL-8B} & \textbf{65.40} & \textbf{75.10} & \textbf{67.41} & \textbf{43.80} & \textbf{64.56} \\
\bottomrule
\end{tabular}

\vspace{3pt}
\begin{tabular}{@{}lccccc@{}}
\toprule
Model & LVBench & MVBench & TempCompass & \makecell{LongVideo\\Bench} & Avg. \\
\midrule
Video-R1~\cite{videor1} & -- & 64.80 & 73.20 & -- & -- \\
Qwen3-VL-8B & 45.64 & 63.75 & 69.38 & 63.80 & 60.56 \\
\textbf{\ours{}-Qwen3-VL-8B} & \textbf{46.20} & \textbf{64.08} & \textbf{76.76} & \textbf{64.20} & \textbf{63.06} \\
\bottomrule
\end{tabular}
\end{table}

Table~\ref{tab:general} reports results on general video benchmarks. Video-R1 results are taken from the original paper, and unreported entries are marked with ``--''. Although \ours{} is not trained on these benchmarks, \ours{}-Qwen3-VL-8B improves over the original Qwen3-VL-8B on all comparable benchmarks, showing that the learned skill is not limited to the in-domain task. The gain on TempCompass is particularly relevant because it depends on temporal boundaries and fine-grained event order. Gains on MMVU, MLVU, MVBench, Video-MME, VideoMMMU, LVBench, and LongVideoBench indicate transfer to broader video QA and multimodal understanding. Overall, the in-domain results validate direct evidence supervision, while the general results show that this supervision acts as a useful proxy for strengthening video spatiotemporal perception; training dynamics and checkpoint results are shown in Appendix Figure~\ref{fig:training_dynamics}.

\subsection{Ablation studies}

Table~\ref{tab:ablation} analyzes four factors: perturbation dimension, task type, shortcut control, and data composition. Perturbation ablations measure the roles of spatial, temporal, and spatiotemporal signals. Task ablations test the complementarity between Grounding and MCQ. Shortcut-control and data-composition ablations evaluate the robustness of the training signal.

\begin{table}[H]
\centering
\caption{Ablation study.}
\label{tab:ablation}
\small
\begin{tabular}{@{}lcc@{}}
\toprule
Setting & \ours{}-Bench & General Avg. \\
\midrule
\ours{} (full) & \textbf{67.40} & \textbf{63.06} \\
\midrule
\textit{Perturbation dimension} \\
\quad Spatial only & 64.80 & 61.84 \\
\quad Temporal only & 65.20 & 61.97 \\
\quad Spatiotemporal only & 65.60 & 62.30 \\
\midrule
\textit{Task type} \\
\quad Grounding only & 63.40 & 61.40 \\
\quad MCQ only & 62.80 & 61.45 \\
\midrule
\textit{Training strategy} \\
\quad Without shortcut control & 65.00 & 61.70 \\
\midrule
\textit{Data composition} \\
\quad Remove Large-Edit subsets & 64.40 & 62.10 \\
\bottomrule
\end{tabular}
\end{table}

Table~\ref{tab:ablation} shows that the full \ours{} obtains the best results on both \ours{}-Bench and the general average. Keeping only one perturbation dimension consistently reduces performance, indicating that spatial, temporal, and spatiotemporal changes provide complementary training signals. Using only Grounding or only MCQ causes larger drops, showing that continuous localization rewards and closed-set attribute discrimination complement each other. Removing shortcut control or the Large-Edit subsets also degrades results, suggesting that randomization, overlap control, and hard edit samples improve the robustness of cross-video difference understanding.

\section{Conclusion}
\label{sec:conclusion}

We present \ours{}, a cross-video difference training framework for Video MLLMs. It creates controlled local differences in real videos and converts A/B video comparison into two verifiable tasks: Grounding and MCQ. This teaches the model to identify changes and localize temporal and spatial evidence. Experiments show that \ours{}-Qwen3-VL-8B achieves stable gains on our cross-video difference benchmark and on multiple general video understanding benchmarks. This suggests that cross-video differences are an effective proxy signal for improving fine-grained spatiotemporal perception. A current limitation is that the differences are still programmatic, and future work can introduce more natural and semantic video changes.

\clearpage
\bibliographystyle{plainnat}
\bibliography{deltavid}

\clearpage
\appendix
\section{Appendix}

\subsection{Source video curation and pre-processing}
\label{app:data}

\trainset{} uses high-quality web videos that we manually search and curate for cross-video spot-the-difference construction. We keep clips with clear visual content, stable subjects, continuous motion, and localizable regions, and remove clips that are too short, heavily compressed, strongly shaky, or visually ambiguous. To avoid leakage from repeated sources, each source video contributes at most one A/B pair. The training set and \ours{}-Bench use disjoint source videos, and we avoid known video clips from public evaluation benchmarks.

\begin{table}[h]
\centering
\caption{Source-video pre-processing.}
\label{tab:app-source}
\small
\begin{tabular}{@{}ll@{}}
\toprule
Item & Setting \\
\midrule
Source videos & Curated high-quality public web videos \\
Source duration & Candidates that support 3--60 second clips \\
Per source video & At most one A/B pair \\
Clip length & 3--60 seconds \\
Frame rate and resolution & 24 FPS, 1280 $\times$ 720, aspect ratio preserved \\
Encoding and audio & H.264, audio removed \\
\bottomrule
\end{tabular}
\end{table}

After pre-processing, Video A keeps the original clip, and Video B adds one or more controlled differences to the same clip. The underlying pool contains 8,200 A/B video pairs and 10,743 difference instances, covering both single-difference and multi-difference cases. Multi-difference samples contain 1--3 differences and limit spatial overlap. When temporal and spatial edits appear in the same sample, the temporal edit is applied first. This avoids label mismatch after the video length changes.

The final training set has 9,433 task samples. Among them, 7,433 samples are paired-video samples. Another 2,000 samples are paired-image spot-the-difference samples extracted from video pairs. These image samples are used only for spatial localization and give direct bbox rewards. The task-level composition is:

\begin{table}[h]
\centering
\caption{Task-level composition of \trainset{}.}
\label{tab:app-taskcomp}
\small
\begin{tabular}{@{}llr@{}}
\toprule
Family & Subtask & Count \\
\midrule
\multirow{2}{*}{Grounding} & Spatial localization (S-IoU) & 3400 \\
& Temporal localization (T-IoU) & 1933 \\
\midrule
\multirow{3}{*}{MCQ} & Shape choice & 400 \\
& Time choice & 2500 \\
& Location choice & 1200 \\
\midrule
\textbf{Total} & -- & \textbf{9433} \\
\bottomrule
\end{tabular}
\end{table}

The detailed training subset distribution is reported in Table~\ref{tab:data}.

\begin{table}[h]
\centering
\caption{\trainset{} training data distribution.}
\label{tab:data}
\small
\begin{tabular}{@{}lrrr@{}}
\toprule
Subset & MCQ & Grounding & Total \\
\midrule
Spatial-Single & 855 & 1476 & 2331 \\
Temporal-Single & 571 & 1500 & 2071 \\
Mixed-Difference & 1006 & 788 & 1794 \\
Dual-Difference & 614 & 569 & 1183 \\
Spatiotemporal-Single & 399 & 0 & 399 \\
Large-Edit/region\_blur & 188 & 247 & 435 \\
Large-Edit/region\_pixelate & 180 & 254 & 434 \\
Large-Edit/color\_shift & 176 & 252 & 428 \\
Large-Edit/add\_shape & 111 & 247 & 358 \\
\midrule
\textbf{Total} & \textbf{4100} & \textbf{5333} & \textbf{9433} \\
\bottomrule
\end{tabular}
\end{table}

\subsection{Perturbation sampling and multi-difference construction}
\label{app:perturbation}

Table~\ref{tab:perturbation} lists all 13 perturbation types. In implementation, perturbation parameters are randomly sampled from preset ranges. These parameters include region position, duration, shape size, hue shift, and speed factor. Typical settings are:

\begin{itemize}[leftmargin=1.5em,itemsep=1pt]
\item \textbf{Time window for spatial perturbations}: covers 15\%--70\% of the video length, with a random start frame.
\item \textbf{Shape size}: 6\%--18\% of the shorter image side; opacity, rotation, and aspect ratio are randomized.
\item \textbf{Local region size}: blur, pixelation, and color-shift regions usually cover 10\%--25\% of frame width and height.
\item \textbf{Hue shift}: sampled from \{40,60,80,100,120\} degrees.
\item \textbf{Blur and pixelation}: blur uses Gaussian blur. Pixelation downsamples the local region and restores it with nearest-neighbor interpolation.
\item \textbf{Temporal edits}: speed factors are sampled from \{0.5x, 1.5x, 2.0x\}. Loop counts are sampled from \{2,3\}. Freeze duration is about 8--24 frames. Segment deletion uses the Video-A timeline; other temporal edits use the Video-B timeline by default.
\item \textbf{Spatiotemporal edits}: object appearance, disappearance, motion, and timed local blur store both time spans and boxes. For moving objects, the box is the union of per-frame boxes over the trajectory.
\end{itemize}

For multi-difference samples, the reward chooses the metric according to each ground-truth difference: spatial differences use S-IoU, temporal differences use T-IoU, and spatiotemporal differences use the product of T-IoU and S-IoU. Matching uses a greedy one-to-one rule: each ground-truth difference selects the highest-scoring unmatched prediction. Extra predictions are not penalized directly, but missing matches reduce the mean score.

\subsection{Shape and color space}
\label{app:appearance}

Shape insertion, object appearance, object disappearance, and moving-object perturbations share the same appearance space. It contains 19 shapes: \texttt{rectangle}, \texttt{circle}, \texttt{triangle}, \texttt{star}, \texttt{diamond}, \texttt{cross}, \texttt{ellipse}, \texttt{pentagon}, \texttt{hexagon}, \texttt{arrow\_up}, \texttt{arrow\_right}, \texttt{heart}, \texttt{crescent}, \texttt{plus}, \texttt{ring}, \texttt{semicircle}, \texttt{trapezoid}, \texttt{parallelogram}, and \texttt{octagon}.

It also contains 24 colors: \texttt{red}, \texttt{green}, \texttt{blue}, \texttt{yellow}, \texttt{cyan}, \texttt{magenta}, \texttt{orange}, \texttt{white}, \texttt{pink}, \texttt{lime}, \texttt{teal}, \texttt{purple}, \texttt{brown}, \texttt{navy}, \texttt{coral}, \texttt{gold}, \texttt{salmon}, \texttt{turquoise}, \texttt{violet}, \texttt{olive}, \texttt{maroon}, \texttt{sky\_blue}, \texttt{spring\_green}, and \texttt{hot\_pink}. The 19 shapes and 24 colors form 456 base appearances. During generation, opacity, rotation, and aspect ratio are also randomized. Thus, the same shape-color pair can still look different across samples.

\subsection{Reward parsing and hard options}
\label{app:reward}

The final reward combines task reward and format reward:
\[
R=(1-\alpha)R_{\mathrm{task}}+\alpha R_{\mathrm{format}},\quad \alpha=0.1.
\]
MCQ uses exact match. Spatial localization uses S-IoU. Temporal localization uses T-IoU. Spatiotemporal localization uses the product of T-IoU and S-IoU. The format reward requires the answer to be parseable from \texttt{<answer>...</answer>}. Grounding samples further require at least one valid \texttt{<diff>} field, and MCQ samples must parse to one option letter. Parseable answers without an explicit \texttt{<think>} field receive only partial format credit.

In Grounding outputs, bounding boxes use normalized integer coordinates in the range $0$--$1000$, and time spans are expressed in seconds. The parser reads only the fields required by the task type: boxes for spatial samples, time spans for temporal samples, and both for spatiotemporal samples. Thus, pure temporal samples are not required to predict meaningless spatial boxes.

MCQ distractors use visually similar shape and color neighbors. For example, neighbors of \texttt{circle} include \texttt{ellipse}, \texttt{ring}, and \texttt{semicircle}; neighbors of \texttt{red} include \texttt{coral}, \texttt{salmon}, \texttt{hot\_pink}, \texttt{maroon}, and \texttt{orange}. Temporal options use coarse intervals, such as the beginning, middle, end, full video, beginning-to-middle, and middle-to-end. Location options come from a 3$\times$3 grid. Options are shuffled and re-lettered after generation.

\subsection{\ours{}-Bench construction and human verification}
\label{app:bench}

\ours{}-Bench is constructed through candidate curation, automatic filtering, and human verification. We first curate about 4,000 candidate A/B samples from carefully selected high-quality video pairs and increase the proportion of multi-difference and spatiotemporal cases to cover diagnostic questions such as difference counting and first-change recognition. We then filter candidates by visibility, temporal length, spatial ambiguity, distance between changes, and distinguishability under reference models.

After automatic filtering, an annotation team checks whether each question is answerable from the videos alone, unambiguous, and consistent with the generated key. A question enters the final set only when the review decision and answer agree. Cases needing correction are reviewed again, and rejected cases are discarded. The keep/reject agreement is $\kappa=0.81$, and answer exact-match agreement among kept samples is 94.6\%. The final 500 questions are sampled to match the skill quotas in Table~\ref{tab:benchcomp}, with a fixed order for reproducible sharding.

\begin{table}[h]
\centering
\caption{\ours{}-Bench diagnostic composition.}
\label{tab:benchcomp}
\small
\begin{tabular}{@{}lrl@{}}
\toprule
Category & Count & Skill \\
\midrule
Difference count & 100 & Count distinct differences in the video pair \\
First change type & 75 & Identify the first change in multi-difference samples \\
Color recognition & 75 & Identify the color of the inserted or changed pattern \\
Region location & 75 & Identify where the difference appears in the frame \\
Shape recognition & 75 & Identify the shape that appears, disappears, or moves \\
Color generalization & 50 & Recognize fine-grained color changes \\
Change-type generalization & 50 & Recognize change types in new combinations \\
\midrule
\textbf{Total} & \textbf{500} & -- \\
\bottomrule
\end{tabular}
\end{table}

\subsection{Training, inference, and compute}
\label{app:training}

\begin{algorithm}[h]
\caption{\ours{} rule-reward post-training pipeline}
\label{alg:training}
\small
\begin{algorithmic}[1]
\Require Real video set $\mathcal{V}$, base model $\pi_\theta$
\Ensure Post-trained model $\pi_{\theta'}$
\State \textbf{// Data construction and task conversion}
\For{$v \in \mathcal{V}$}
  \State Sample a perturbation rule $p$ and create paired video $v'$
  \State Record difference labels and task metadata
  \State Convert labels into Grounding and MCQ samples and add them to $\mathcal{D}$
\EndFor
\State \textbf{// Direct GRPO post-training}
\For{each training step}
  \State Sample a task sample $(x, q, y^\star)$ from $\mathcal{D}$ and generate $G$ answers
  \State Compute rule-based task rewards $\mathbf{r}$ by task type
  \State Update model parameters $\theta$ with GRPO
\EndFor
\State \Return $\pi_{\theta'} \leftarrow \pi_\theta$
\end{algorithmic}
\end{algorithm}

\begin{table}[h]
\centering
\caption{Main training configuration.}
\label{tab:app-traincfg}
\small
\begin{tabular}{@{}ll@{}}
\toprule
Item & Setting \\
\midrule
Base model & Qwen3-VL-8B \\
Trainable parameters & LoRA, rank 64, $\alpha=128$, dropout 0.05 \\
Precision and parallelism & bfloat16, DeepSpeed ZeRO-2 \\
Generations per sample & $G=8$ \\
Learning rate and KL & $1\times10^{-6}$, $\beta=0.02$ \\
Optimizer & AdamW \\
Max video frames & Up to 60 during training; 32 or 64 for evaluation \\
Effective batch size & 64 \\
Hardware & 1 node with 8$\times$H800-80GB GPUs \\
\bottomrule
\end{tabular}
\end{table}

On \ours{}-Bench, all reference models use the same videos, questions, answer parser, and 32 uniformly sampled frames. General video benchmarks use a unified evaluation harness; image-style or short MCQ benchmarks use 32 frames, and video understanding benchmarks use 64 frames. Open-ended MMVU/MLVU subsets use the standard VLMEvalKit GPT-judge setting without modifying the default judge prompt.

\begin{table}[h]
\centering
\caption{Approximate compute budget.}
\label{tab:app-compute}
\small
\begin{tabular}{@{}lr@{}}
\toprule
Stage & Cost \\
\midrule
A/B pair generation & about 600 CPU hours \\
One GRPO post-training run & about 24 hours on 8$\times$H800 GPUs ($\approx$192 H800 GPU hours) \\
\ours{}-Bench inference & about 40 GPU hours \\
General video benchmark evaluation & about 220 GPU hours \\
Reference-baseline inference & about 18 GPU hours \\
\midrule
\textbf{Total excluding CPU data generation} & \textbf{about 470 GPU hours} \\
\bottomrule
\end{tabular}
\end{table}

\subsection{Training dynamics and checkpoint performance}
\label{app:training_dynamics}

\begin{figure}[t]
\centering
\begin{subfigure}{0.49\textwidth}
  \centering
  \includegraphics[width=\linewidth]{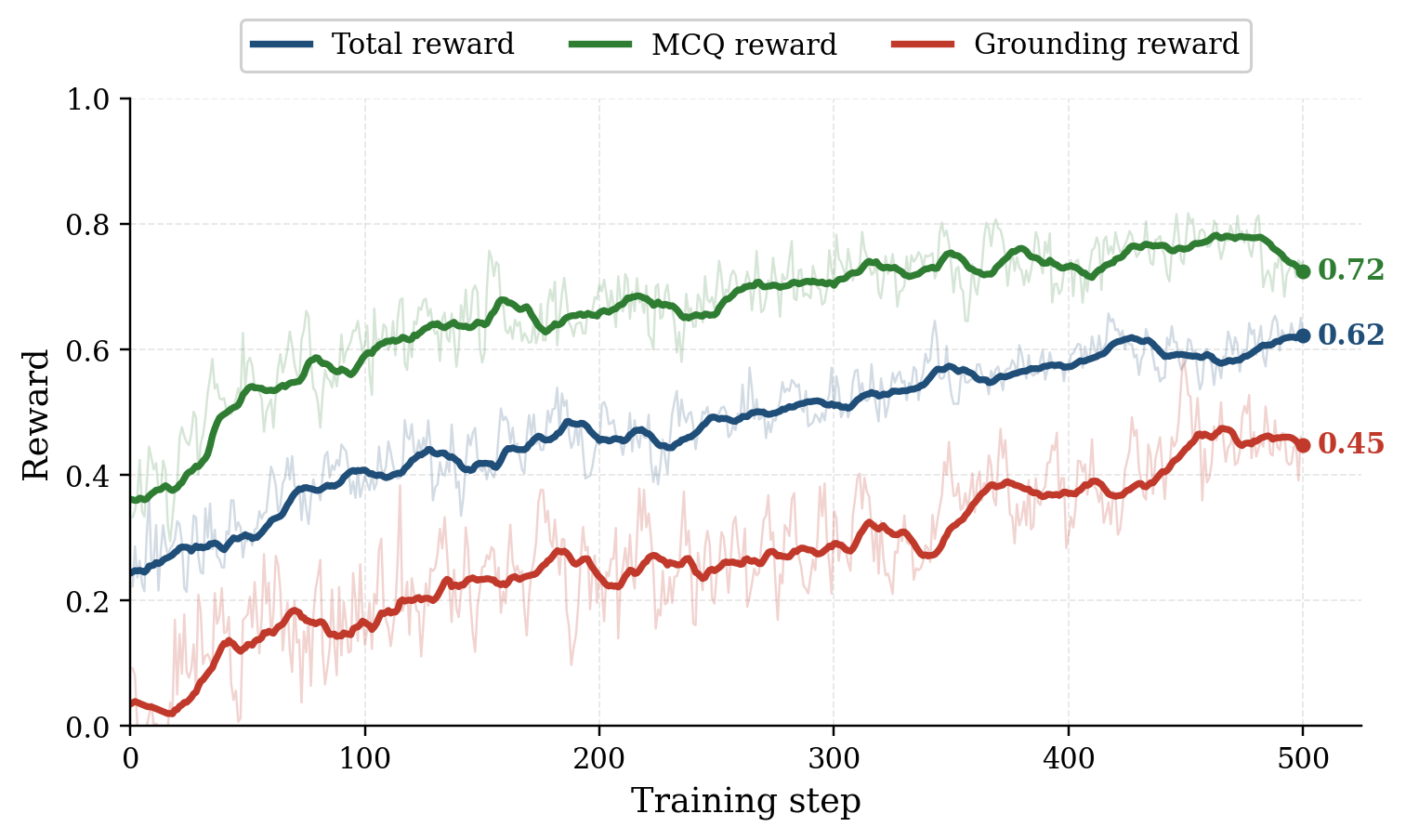}
  \caption{Rule-reward training curves.}
  \label{fig:reward_curve}
\end{subfigure}
\hfill
\begin{subfigure}{0.49\textwidth}
  \centering
  \includegraphics[width=\linewidth]{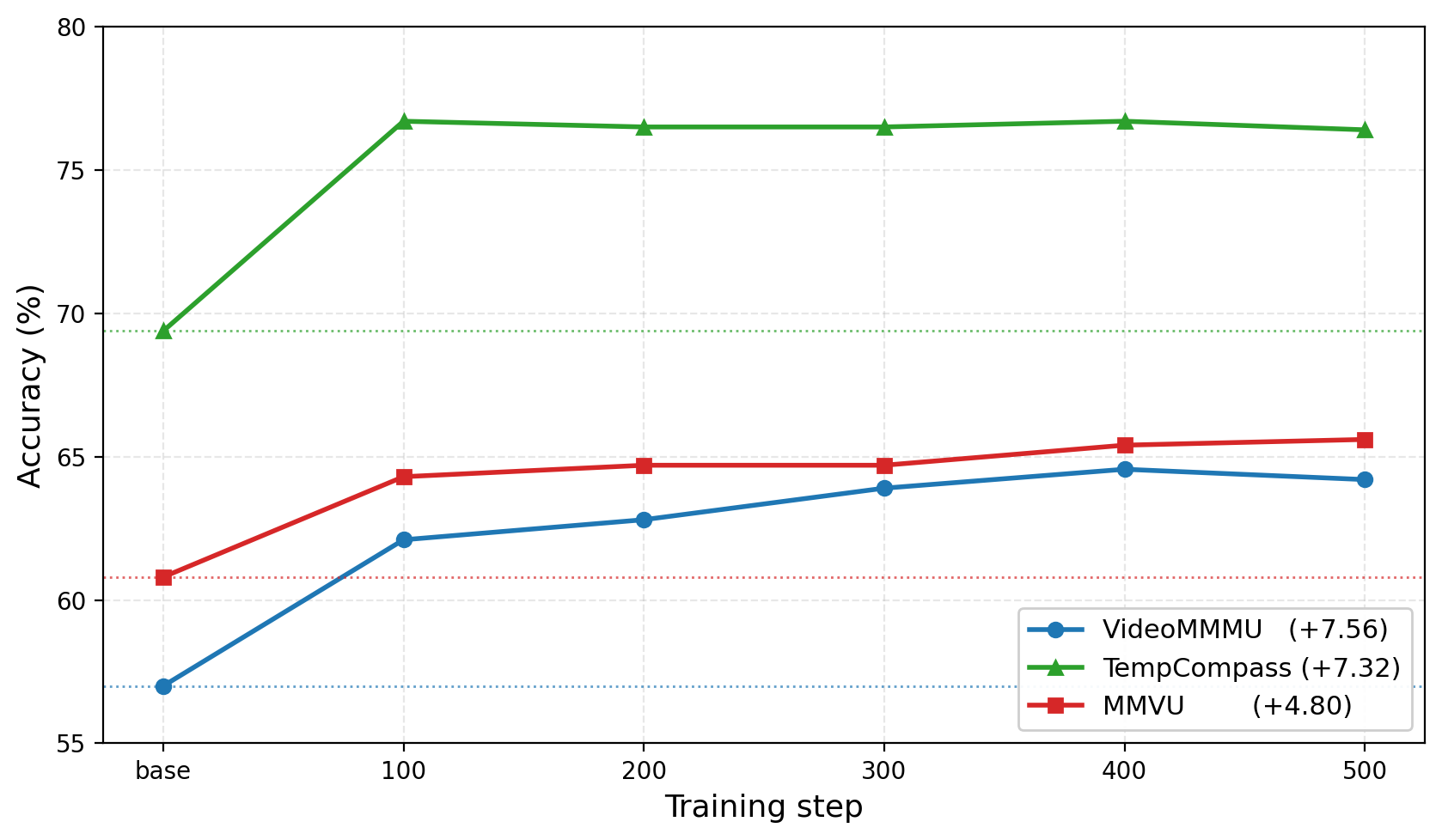}
  \caption{Checkpoint performance on representative benchmarks.}
  \label{fig:ckpt_benchmarks}
\end{subfigure}
\caption{Training dynamics and checkpoint performance. \textbf{Left}: MCQ reward, Grounding reward, and total reward during GRPO post-training. \textbf{Right}: accuracy of different checkpoints on VideoMMMU, TempCompass, and MMVU; dashed lines indicate base-model performance.}
\label{fig:training_dynamics}
\end{figure}

Figure~\ref{fig:training_dynamics} shows the reward trajectory and checkpoint performance during GRPO post-training. The left panel shows that MCQ reward, Grounding reward, and total reward steadily improve as training progresses; the right panel reports checkpoint accuracy on VideoMMMU, TempCompass, and MMVU to track whether reward improvements are accompanied by changes on general video benchmarks.

\subsection{MCQ templates and answer format}
\label{app:mcq}

This section gives representative MCQ templates. The main training uses shape, time, and location questions. Color and change-type templates are mainly used for independent diagnostic evaluation. To avoid ambiguity, each answer is a single option letter. In real samples, time, location, number of options, and distractors are randomly generated by the program. The format reward requires the final answer to be parseable from \texttt{<answer>...</answer>}.

\mypara{Template A: change type (diagnostic evaluation).}
\begin{quote}
\small
What is the difference between video A and video B?\\
(A) During 2.1s--5.3s, the upper-right region of video B is blurred.\\
(B) During 2.1s--5.3s, the lower-left region of video B is blurred.\\
(C) There is no difference.\\
(D) During 2.1s--5.3s, this segment of video B is played at 2.0x speed.\\
Output: \texttt{<answer>A</answer>}
\end{quote}

\mypara{Template B: change location.}
\begin{quote}
\small
Where does the difference appear in the frame?\\
(A) Upper-left corner\\
(B) Center\\
(C) Lower-right corner\\
(D) Middle-left area\\
Output: \texttt{<answer>C</answer>}
\end{quote}

\subsection{Grounding output format}
\label{app:grounding}

The Grounding task asks the model to output structured evidence that matches the perturbation dimension. The following examples show the key fields used by the reward module. \texttt{time} is the start and end time in seconds. \texttt{bbox} is the normalized bounding box $[x_1, y_1, x_2, y_2]$, with coordinates from 0 to 1000. The reward module only depends on the time span and/or spatial box fields. It does not depend on fixed wording or extra descriptions.

\begin{quote}
\small
Spatial perturbation: \texttt{\{bbox: [650,100,950,450]\}}\\
Temporal perturbation: \texttt{\{time: [2.1, 5.3]\}}\\
Spatiotemporal perturbation: \texttt{\{time: [2.1, 5.3], bbox: [650,100,950,450]\}}
\end{quote}

For spatial perturbation samples, the prompt gives the time window, and the reward mainly computes bbox IoU. For temporal perturbation samples, the reward computes temporal IoU. For spatiotemporal samples, the reward computes the product of temporal IoU and spatial IoU. The model may output explanatory reasoning, but scoring uses the parseable structured fields in \texttt{<answer>...</answer>}.

\subsection{Statistical checks, release safeguards, and annotation details}
\label{app:responsible}

\mypara{Statistical checks.}
For the main benchmark comparisons, evaluation is performed on fixed test sets with the same input frames, prompts, answer parser, and scoring rules for all models. We use item-level resampling over evaluation examples to check whether the observed trends are stable for the reported aggregate scores. The conclusions in the main tables remain unchanged under these checks.

\mypara{Broader impacts and safeguards.}
\ours{} is intended as a research framework for diagnosing and improving fine-grained video understanding. Potential benefits include better temporal grounding, local evidence attribution, and more reliable video QA. Potential risks include applying fine-grained change detection to privacy-sensitive monitoring or surveillance scenarios. To reduce such risks, released data and checkpoints should be accompanied by research-use terms, unsafe-content filtering, removal of personal identifiers when applicable, and documentation of intended and out-of-scope uses.

\mypara{Licenses and existing assets.}
All existing models, datasets, and benchmarks used in the paper are credited through citations. The release documentation lists the source, version or access point, and license or terms of use when available. Derived assets should preserve the applicable attribution requirements and should not redistribute restricted content beyond the permissions of the original source.

\mypara{Human verification.}
Human verification is used only for dataset quality control. Annotators are instructed to check whether each example is answerable from the provided videos, whether the question is unambiguous, and whether the generated answer matches the visible evidence. The annotation package includes written instructions, interface examples, and compensation information; no personal or sensitive information is collected from annotators or video subjects for the purpose of this verification.

\end{document}